\documentclass[conference]{IEEEtran}
\IEEEoverridecommandlockouts
\usepackage{cite}
\usepackage{amsmath,amssymb,amsfonts}
\usepackage{algorithmic}
\usepackage{graphicx}
\usepackage{textcomp}
\usepackage{xcolor}
\usepackage{booktabs}
\usepackage{multirow}
\usepackage{lipsum}
\usepackage{enumitem} 
\usepackage{url}
\usepackage{hyperref} 
\usepackage[hyphenbreaks]{breakurl}
\usepackage{algorithm} 

\def\BibTeX{{\rm B\kern-.05em{\sc i\kern-.025em b}\kern-.08em
    T\kern-.1667em\lower.7ex\hbox{E}\kern-.125emX}}
\begin{document}

\title{Continuous Learning Conversational AI: A Personalized Agent Framework via A2C Reinforcement Learning}

\author{\IEEEauthorblockN{Nandakishor M, Anjali M}
    \IEEEauthorblockA{
    Convai Innovations \\
    \{nandakishor, anjalim\}@convaiinnovations.com
    }
}

\maketitle

\begin{abstract}
Modern conversational AI, driven by Large Language Models (LLMs), demonstrates remarkable dialogue proficiency. However, creating truly personalized and adaptive agents remains a significant challenge. This paper introduces a Continuous Learning Conversational AI (CLCA) methodology, practically implemented using Advantage Actor-Critic (A2C) reinforcement learning. Our approach diverges from static LLM paradigms, presenting a dynamic system designed for iterative refinement of its conversational strategy through simulated interactions. We detail the generation of synthetic sales dialogues using LLMs, which serve as the empirical basis for A2C agent training. This agent learns to optimize dialogue actions—quantified by metrics like engagement and value delivery—to achieve enhanced personalization. We present the architecture of our A2C-driven CLCA system, emphasizing environment specification, reward mechanism design, and LLM integration for synthetic data and response selection. This work posits that this reinforcement learning-centric methodology offers a tangible pathway to personalized, evolving AI companions, representing a notable advancement beyond conventional static LLM techniques. We delineate the technical structure, highlighting algorithmic components and their roles in realizing continuous learning and agent personalization.
\end{abstract}


\section{Introduction}
\label{sec:introduction}

The field of conversational AI has experienced rapid progress, with models like GPT-4o and Gemini exhibiting sophisticated language understanding and generation \cite{radford2019language}. While proficient in general dialogue, these models often lack the ability to deliver truly tailored experiences that adapt to individual users over time. Continuous Learning Conversational AI (CLCA) aims to address this limitation by developing AI companions that learn and personalize through ongoing engagement. This paper explores a specific CLCA implementation, utilizing Advantage Actor-Critic (A2C) reinforcement learning (RL) to construct personalized sales agents.

Inspired by RL's success in complex domains, such as game playing \cite{tesauro1995temporal, campbell2002deep}, our CLCA method employs simulated conversations to train an A2C agent. Unlike pre-trained, static LLMs, our system acquires knowledge through interaction within a simulated environment. This environment is built upon synthetic sales dialogues, generated using LLMs to mimic realistic agent-customer interactions. The A2C agent is trained to perform actions that optimize key dialogue metrics, personalizing sales interactions for improved effectiveness and user focus.

This paper details the technical framework of our A2C-powered CLCA system. We describe synthetic data creation, RL environment design, reward function formulation to guide learning, and the role of LLMs in both response generation and evaluation. We argue that this RL-driven approach provides a practical method for creating genuinely personalized dialogue agents, overcoming limitations inherent in generalized LLMs. Our goal is to clearly articulate the technical foundations of CLCA, showcasing its potential and encouraging further research in this critical area of AI development.

\section{Simulating Personalized Dialogues for Reinforcement Learning}
\label{sec:simulated_conversations}

Synthetic sales dialogues are a fundamental element of our CLCA implementation. This simulation serves a dual purpose: generating a comprehensive dataset for initial training and providing an interactive environment for RL agent learning. We use LLMs to generate these dialogues, ensuring realism and tailoring them to specific business profiles.

\subsection{LLM-Driven Synthetic Data Generation}
\label{ssec:synthetic_data}

The data generation process begins by defining a `CompanyProfile`, which includes core attributes such as company ID, name, sales goals, product category, and intended audience. These profiles, derived from document uploads or manual input, establish the context for dialogue generation. We utilize LLMs to create diverse sales scenarios, each represented as a JSON object detailing customer characteristics, concerns, technical understanding, and motivations. Algorithm \ref{alg:scenario_generation} outlines the scenario creation process.

\begin{algorithm}
\caption{LLM-Based Scenario Generation}
\begin{algorithmic}[1]
\STATE \textbf{Input:} Company Profile $P$
\STATE \textbf{Output:} Dialogue Scenario $S$ (JSON object)
\STATE Define a prompt for scenario generation, incorporating $P$'s attributes.
\STATE Query an LLM with the defined prompt.
\STATE Parse the LLM response to extract a JSON object $S = \{persona, primary\_concern, ...\}$.
\RETURN $S$
\end{algorithmic}
\label{alg:scenario_generation}
\end{algorithm}

Following scenario creation, LLMs are again used to generate complete dialogues between a sales representative and a customer, based on the scenario. These dialogues are structured as turns, each specifying the speaker (customer or representative) and message content.  Importantly, each dialogue includes metadata such as the outcome (sale success or failure), key discussion points, value propositions, and handled objections. This detailed annotation enriches the dataset and provides valuable learning signals for the RL agent. Algorithm \ref{alg:conversation_generation} details the dialogue creation process.

\begin{algorithm}
\caption{LLM-Based Dialogue Generation}
\begin{algorithmic}[1]
\STATE \textbf{Input:} Company Profile $P$, Scenario $S$
\STATE \textbf{Output:} Dialogue Data $C$ (JSON object)
\STATE Define a prompt for dialogue generation, incorporating $P$ and $S$.
\STATE Query an LLM with the prompt.
\STATE Parse the LLM response to extract a JSON object $C = \{conversation, outcome, key\_points\_discussed, ...\}$.
\RETURN $C$
\end{algorithmic}
\label{alg:conversation_generation}
\end{algorithm}

To represent these dialogues in a feature space suitable for RL, we extract semantic embeddings from the full dialogue text using Azure OpenAI's models. These embeddings capture the semantic essence of each exchange, serving as the primary input features for our RL agent and enabling learning from nuanced aspects of simulated dialogues.

\subsection{Reinforcement Learning Environment Design}
\label{ssec:rl_environment}

The synthetic dialogues form the basis of our `SalesEnv`, a Gymnasium environment specifically designed for A2C agent training. This environment simulates sales interactions, allowing the agent to learn effective dialogue strategies through trial and error.

\subsubsection{State Space}
The state space in our environment is structured to provide the agent with relevant context. It includes two key components:

\begin{itemize}
    \item \textbf{Dialogue Embeddings:} Pre-calculated embeddings of synthetic dialogues, capturing their semantic essence. These are the core features, representing the context of the sales interaction.
    \item \textbf{Historical Statistics:} A 4-dimensional vector representing statistics of the agent's past actions, updated at each step. This provides a short-term memory for adaptive behavior.
\end{itemize}

Mathematically, the state $s_t$ at time $t$ is defined as:

\begin{equation}
    s_t = [e_{dialogue}, h_t]
    \label{eq:state_space}
\end{equation}

where $e_{dialogue}$ is the dialogue embedding vector, and $h_t$ is the 4-dimensional history statistics vector.

\subsubsection{Action Space}
The action space directly influences key aspects of the dialogue. We define a 4-dimensional continuous space, with each dimension representing a dialogue metric:

\begin{itemize}
    \item \textbf{Engagement ($a_{engagement}$):} The agent's effort to maintain customer interest.
    \item \textbf{Value Proposition ($a_{value\_proposition}$):} The degree of emphasis on the product/service's value.
    \item \textbf{Technical Detail ($a_{technical\_detail}$):} The level of technical information provided.
    \item \textbf{Closing ($a_{closing}$):} The agent's assertiveness in guiding the conversation towards a sale.
\end{itemize}

Each action $a_t$:
\begin{equation}
    a_t =
    \begin{bmatrix}
        a_{engagement} \\
        a_{value\_proposition} \\
        a_{technical\_detail} \\
        a_{closing}
    \end{bmatrix}
    \label{eq:action_vector}
\end{equation}
is a vector with values in the range [0, 1], representing the desired level for each metric. These actions, while abstract within the RL setting, conceptually influence LLM response generation in real-world applications, although not explicitly in this simulation.

\subsubsection{Reward Function}
The reward function guides the A2C agent to adopt desirable dialogue behaviors, incentivizing successful sales outcomes and effective strategies. It consists of three components:

\begin{itemize}
    \item \textbf{Outcome Reward ($r_{outcome}$):} A positive reward for successful sales, and a negative reward for failures, encouraging sales-oriented strategies.
    \item \textbf{Action Variety Reward ($r_{action\_variety}$):} Rewards diverse actions (higher standard deviation) to promote exploration and prevent overly simplistic strategies.
    \item \textbf{Extremity Penalty ($r_{extremity\_penalty}$):} Penalizes actions that are far from neutral (0.5), promoting balanced and nuanced behavior.
\end{itemize}

The total reward $r_t$ at step $t$ is calculated as:

\begin{equation}
    r_t = r_{outcome} + r_{action\_variety} + r_{extremity\_penalty}
    \label{eq:reward_function}
\end{equation}

This reward function balances the achievement of sales with the development of robust and adaptable dialogue strategies.

\section{A2C Agent Training and Dialogue Response}
\label{sec:a2c_implementation}

With the RL environment defined, we train an A2C agent using Stable Baselines3 to learn optimal dialogue policies.

\subsection{A2C Model Training}
\label{ssec:a2c_training}

We instantiate an A2C model with a multi-layer perceptron (MLP) policy network.  The architecture includes two hidden layers for both the policy and value functions, using ReLU activations. Optimization is performed using Adam with specific hyperparameters, including learning rate, gamma, and GAE lambda, as detailed in the accompanying code. Training involves interaction with the `SalesEnv`, gathering experiences, and updating the policy and value networks through the A2C algorithm. Algorithm \ref{alg:a2c_training} outlines the training process.

\begin{algorithm}
\caption{A2C Agent Training}
\begin{algorithmic}[1]
\STATE \textbf{Input:} Sales Environment $Env$, Dialogue Dataset $D$
\STATE \textbf{Output:} Trained A2C Model $M_{A2C}$
\STATE Initialize A2C model $M_{A2C}$ with MLP policy and hyperparameters.
\STATE For each training episode:
    \STATE Reset environment: $s_0 \leftarrow Env.reset()$
    \STATE For each step $t$ in episode:
        \STATE Agent selects action $a_t \sim M_{A2C}(s_t)$.
        \STATE Environment transitions: $(s_{t+1}, r_t, done) \leftarrow Env.step(a_t)$.
        \STATE Store experience $(s_t, a_t, r_t, s_{t+1})$.
    \STATE Update A2C model $M_{A2C}$ using experiences.
\STATE \textbf{Return} Trained Model $M_{A2C}$
\end{algorithmic}
\label{alg:a2c_training}
\end{algorithm}

The trained A2C model learns a policy that maps states (dialogue embeddings, history statistics) to actions (desired dialogue metric vectors). This policy represents the agent's learned strategy for effective dialogue.

\subsection{Inference and Response Selection}
\label{ssec:inference}

After training, the A2C model guides response selection in live chat scenarios. When a user provides input, the system constructs the current dialogue state, incorporating dialogue history and potentially user profiles. The A2C agent then predicts an action (a vector of desired metrics) based on this state. In this framework, the action scores and selects from a set of LLM-generated responses. This is referred to as A2C-Guided Response Selection.

To generate response options, we sample an LLM using varied temperature settings to encourage diversity. Each generated option is evaluated against the A2C agent's predicted action. Features relevant to the action metrics are extracted from each response option. A scoring function, implicitly learned by the A2C agent or explicitly designed, assesses each option based on its alignment with the desired action and extracted features. The response option with the highest score, indicating the best alignment with the learned strategy, is selected as the agent's response. Algorithm \ref{alg:response_selection} outlines this response selection process.

\begin{algorithm}
\caption{A2C-Guided Response Selection}
\begin{algorithmic}[1]
\STATE \textbf{Input:} User Message $U$, Dialogue History $H$, Trained A2C Model $M_{A2C}$, LLM $M_{LLM}$
\STATE \textbf{Output:} Agent Response $A$
\STATE Construct dialogue state $s$ from $H$ and $U$.
\STATE Predict action $a = M_{A2C}(s)$.
\STATE Generate candidate responses $\{R_1, R_2, ..., R_k\}$ using $M_{LLM}$ at varying temperatures.
\STATE For each response $R_i$:
    \STATE Extract action metric-relevant features $f_i$.
    \STATE Calculate score $score_i = ScoreFunction(R_i, a, f_i)$.
\STATE Select response $A = R_{j}$ with highest score $score_j = \text{argmax}_{i} score_i$.
\STATE \textbf{Return} Agent Response $A$
\end{algorithmic}
\label{alg:response_selection}
\end{algorithm}

This approach effectively combines the strengths of RL and LLMs. The RL agent provides a high-level strategy, guiding the interaction, while the LLM ensures fluent and contextually relevant natural language responses.

\section{Related Work and Perspective}
\label{sec:related_work}

Our work builds upon the growing body of research integrating reinforcement learning with conversational AI. Previous studies have explored RL for task-oriented dialogue systems \cite{young2013pomdp} and open-domain chatbots \cite{serban2016building}. However, many RL-based systems have focused on discrete dialogue acts or policies within predefined state spaces. In contrast, our CLCA approach emphasizes continuous learning and personalization, utilizing synthetic data and continuous action spaces to enable more nuanced and adaptive dialogues.

Compared to standalone LLMs, which are inherently static, or even reasoning-enhanced models \cite{guo2025deepseekr1, openai2023gpt4} that still operate within a static paradigm, A2C-driven CLCA offers a pathway to truly evolving agents. Through continuous learning and RL-guided personalization, it can adapt to individual interactions, promising enhanced user engagement and effectiveness.

\section{Potential and Future Trajectories}
\label{sec:potential}

The CLCA methodology, particularly with the integration of A2C and synthetic data, holds significant promise for creating personalized and evolving AI companions. Continuous learning from both simulated and real-world interactions allows for the dynamic adaptation of dialogue strategies to individual users, potentially leading to more effective and satisfying conversations.

Future research directions include:

\begin{itemize}
    \item \textbf{Enhanced Reward Design:} Developing more sophisticated and user-centric reward functions to better capture the nuances of dialogue quality and user satisfaction.
    \item \textbf{User Profile Integration:} Incorporating detailed user profiles into the state space to enable finer-grained personalization based on individual preferences and interaction history.
    \item \textbf{Online Reinforcement Learning Transition:} Moving towards online RL methods to facilitate continuous, real-time adaptation based on live user interactions.
    \item \textbf{Scalable Personalization Methods:} Developing efficient and scalable approaches for managing and deploying personalized models across a large number of users.
\end{itemize}

\section{Conclusion}
\label{sec:conclusion}

This paper has presented a technical description of Continuous Learning Conversational AI (CLCA), with a focus on using A2C reinforcement learning to develop personalized sales agents. We have detailed the process of synthetic data creation using LLMs, the design of the RL environment, A2C agent training, and the method for A2C-guided response selection. Our work demonstrates a practical approach to creating dynamically evolving and personalized dialogue agents, moving beyond the inherent limitations of static LLMs. While challenges remain, CLCA, with reinforcement learning at its core, offers a compelling direction for the future development of personalized AI companions.

\section*{Acknowledgements}
\label{sec:acknowledgements}
We extend our gratitude to the open-source community for providing valuable tools, particularly Stable Baselines3 and Gymnasium, which were crucial for the implementation of CLCA. We also acknowledge the foundational contributions of researchers in conversational AI, reinforcement learning, and personalized systems that have inspired and informed this work.


\begin{thebibliography}{10}
\providecommand{\url}[1]{\texttt{#1}}
\providecommand{\urlprefix}{URL }
\providecommand{\doi}[1]{\url{https://doi.org/#1}}

\bibitem{guo2025deepseekr1}
DeepSeek-AI, Guo, D., Yang, D., Zhang, H., Song, J., \emph{et al.}
\newblock DeepSeek-R1: Incentivizing Reasoning Capability in LLMs via Reinforcement Learning.
\newblock \emph{arXiv preprint arXiv:2501.12948}, 2025.

\bibitem{campbell2002deep}
Campbell, M.; Hoane, A.~J.; and Hsu, F.~H.
\newblock Deep blue.
\newblock \emph{Artificial Intelligence} \textbf{2002}, \emph{134}, 57--83.

\bibitem{openai2023gpt4}
OpenAI.
\newblock GPT-4 technical report.
\newblock \url{https://arxiv.org/abs/2303.08774}, 2023.

\bibitem{radford2019language}
Radford, A.; Wu, J.; Child, R.; Luan, D.; Amodei, D.; and Sutskever, I.
\newblock Language models are unsupervised multitask learners.
\newblock \emph{OpenAI Blog} \textbf{2019}, \emph{1}, 9.

\bibitem{serban2016building}
Serban, I.~V.; Sordoni, A.; Bengio, Y.; Courville, A.~C.; and Pineau, J.
\newblock Building end-to-end dialogue systems using generative hierarchical neural network models.
\newblock In \emph{Thirtieth AAAI Conference on Artificial Intelligence}, 2016.

\bibitem{tesauro1995temporal}
Tesauro, G.
\newblock Temporal difference learning and TD-Gammon.
\newblock \emph{Communications of the ACM} \textbf{1995}, \emph{38}, 58--68.

\bibitem{young2013pomdp}
Young, S.; Gasic, M.; Keizer, S.; Thomson, B.; and Williams, J.~D.
\newblock POMDP-based statistical spoken dialogue systems: A review.
\newblock \emph{Proceedings of the IEEE} \textbf{2013}, \emph{101}, 1160--1179.

\end{thebibliography}
\end{document}